\pdfoutput=1

\documentclass[11pt]{article}

\usepackage[preprint]{acl}

\usepackage{times}
\usepackage{latexsym}

\usepackage[T1]{fontenc}

\usepackage[utf8]{inputenc}

\usepackage{microtype}

\usepackage{inconsolata}

\usepackage{graphicx}
\usepackage{fontawesome5}

\usepackage{marvosym}
\usepackage{textcomp} 
\usepackage{amsmath}
\usepackage[section]{placeins}
\usepackage{multirow}
\usepackage{makecell}
\usepackage{amssymb}
\usepackage{xspace}
\usepackage{enumitem}
\usepackage{tcolorbox}
\usepackage{booktabs}
\usepackage{multicol}
\usepackage{cuted}
\usepackage{tabularx}
\usepackage{hyperref}
\usepackage{listings}
\usepackage{mathptmx}
\usepackage{xcolor}
\usepackage{subfigure}
\usepackage{color}
\usepackage{algorithm}
\usepackage{algorithmic}
\definecolor{codegray}{rgb}{0.5, 0.5, 0.5}
\definecolor{codepurple}{rgb}{0.5, 0, 0.5}
\lstdefinestyle{mystyle}{
    keywordstyle= \color{ blue!70},			
    commentstyle= \color{red!50!green!50!blue!50!},		
    numberstyle=\tiny\color{codegray},		
    stringstyle=\color{codepurple},
    basicstyle=\ttfamily,
    breakatwhitespace=false,         
    breaklines=true,		
    captionpos=b,                    
    keepspaces=true,             
    numbersep=5pt,                  
    showspaces=false,                
    showstringspaces=false,		
    showtabs=false,                  
    tabsize=2,
    frame=shadowbox,	
}

\newcommand{\ignore}[1]{}

\usepackage{lipsum}
\usepackage{environ}
\NewEnviron{SMALL}{%
    \scalebox{1}{$\BODY$} 
} 

\newcommand{\cloudicon}{\ensuremath{%
  \mathchoice{\includegraphics[height=2ex]{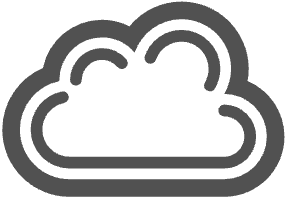}}
    {\includegraphics[height=2ex]{source/cloud.png}}
    {\includegraphics[height=1.5ex]{source/cloud.png}}
    {\includegraphics[height=1ex]{source/cloud.png}}
}}

\definecolor{deepgreen}{RGB}{0, 70, 0}
\definecolor{deepred}{RGB}{182, 32, 22}

\title{\cloudicon AIR: Complex Instruction Generation via Automatic\\ Iterative Refinement}

\usepackage{url} 

\author{
    Wei Liu\textsuperscript{*}, 
    Yancheng He\textsuperscript{*}, 
    Hui Huang\textsuperscript{*,\dag}, 
    Chengwei Hu, 
    Jiaheng Liu, \\[8pt]
    \textbf{Shilong Li},
    \textbf{Wenbo Su}, 
    \textbf{Bo Zheng}
    \\[5pt]
    Alibaba Group \\[3pt]
    \texttt{\{lw02131882,hh456524\}@alibaba-inc.com}
}

\begin{document}
\maketitle

\let\oldthefootnote\thefootnote
\renewcommand{\thefootnote}{} 
\footnotetext{$^{*}$ Equal contribution. $^{\dag}$ Corresponding Author.}
\footnotetext{\textsuperscript{1}Codes are openly available at \url{https://github.com/WeiLiuAH/AIR-Automatic-Iterative-Refinement}.}
\let\thefootnote\oldthefootnote 

\begin{abstract}
With the development of large language models, their ability to follow simple instructions has significantly improved. However, adhering to complex instructions remains a major challenge. Current approaches to generating complex instructions are often irrelevant to the current instruction requirements or suffer from limited scalability and diversity. Moreover, methods such as back-translation, while effective for simple instruction generation, fail to leverage the rich knowledge and formatting in human written documents.
In this paper, we propose a novel \textbf{A}utomatic \textbf{I}terative \textbf{R}efinement  (\textbf{AIR}) framework to generate complex instructions with constraints,
which not only better reflects the requirements of real scenarios but also significantly enhances LLMs' ability to follow complex instructions.
The AIR framework consists of two stages: 1) Generate an initial instruction from a document; 2) Iteratively refine instructions with LLM-as-judge guidance by comparing the model's output with the document to incorporate valuable constraints.
Finally, we construct the AIR-10K dataset with 10K complex instructions and demonstrate that instructions generated with our approach significantly improve the model’s ability to follow complex instructions, outperforming existing methods for instruction generation\textsuperscript{1}.

\end{abstract}
\section{Introduction}

\begin{figure}[t]
    \centering
    \includegraphics[width=1\linewidth]{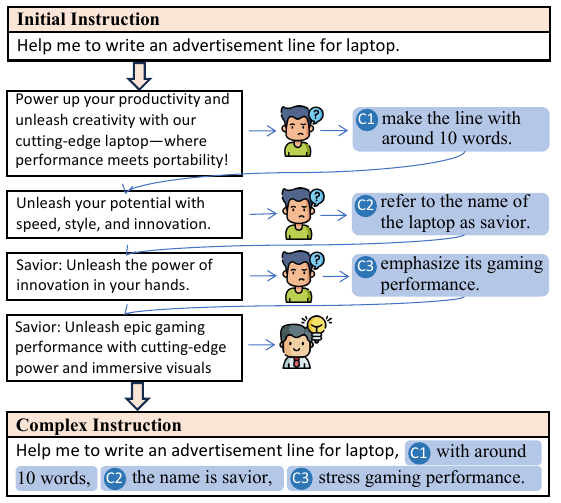}
    \caption{Illustration of how humans iteratively refine instructions to be more complex.}
    \label{fig: intro}
    \vspace{-1mm}
\end{figure}

Recent advancements in Large Language Models (LLMs) have shown impressive performance across a wide range of tasks~\cite{zhao2023survey, li2024graphreader, he2024chinesesimpleqa}. Driven by vast amounts of data and efficient training, most current LLMs are capable of effectively following user instructions and aligning to a certain extent with human preferences~\cite{ouyang2022training,li20242d,huang2025musc}.
However, despite these successes, they still face significant challenges when it comes to following complex instructions~\cite{jiang2023followbench,wen2024benchmarking}.

Existing complex instructions datasets primarily originate from two sources: 1) Curated data from open-source datasets or human annotations~\cite{zhou2024lima,zhang2024cfbench}, which are resource-intensive and \textbf{lack scalability}, and 2) Transforming simple instructions into complex ones automatically using proprietary LLMs~\cite{xu2023wizardlm,sun2024conifer}. 
While the automatic transformation improves scalability, the generated constraints are often recombinations of few-shot examples, resulting in \textbf{limited diversity}.
Moreover, these constraints may have \textbf{low relevance} with the target output, failing to reflect real-world scenarios.



Recently, back-translation, which involves translating text from the target side back into the source side, has been proposed to generate scalable and diverse instructions from human-written corpora~\cite{sennrich2015improving,hoang2018iterative,zheng2024kun,li2023self}. 
However, these methods typically focus on generating \textbf{simple instructions} and have not fully explored the rich knowledge contained in the human corpus.

In this paper, we propose an \textbf{Automatic Iterative Refinement (AIR)} framework for generating high-quality complex instructions.
Specifically, our approach is based on two key observations. First, human-written documents contain massive human preferences that can be converted to specific constraints, such as formatting conventions in legal documents. Second, human often refine complex instructions iteratively based on feedback from model outputs. As illustrated in Figure~\ref{fig: intro}, simple instructions are progressively adjusted and enriched to better align with human preferences. This iterative process plays a critical role in crafting effective complex instructions.

Therefore, our AIR framework incorporates document-based knowledge and LLM-as-judge to iteratively construct complex instructions. 
The framework consists of two key steps: 1) \textbf{Initial Instruction Generation}, where the model generates initial instructions based on the document content; 2) \textbf{Iterative Instruction Refinement}, where instructions are iteratively refined with LLM-as-judge guidance by comparing model outputs with the document, to identify and incorporate valuable constraints. This process enables the framework to generate more challenging instructions that align more closely with real-world scenarios. 

In summary, our contributions are as follows: 

\begin{itemize}[leftmargin=4mm]
    \item  To better align with real-world scenarios, we propose the \textbf{AIR} framework, which iteratively refines complex instructions with LLM-as-judge guidance by comparing with the document.

    \item We introduce a novel instruction dataset, \textbf{AIR-10K}, generated using our framework. Experimental results demonstrate that our fine-tuned model significantly outperforms existing methods on instruction-following benchmarks.

    \item We provide a comprehensive experimental analysis to evaluate the individual components of our framework, validating the contribution of each step to the overall improvement.
\end{itemize}

\section{Related Work}

\subsection{Instruction Generation}

Instruction tuning is essential for aligning Large Language Models (LLMs) with user intentions~\cite{ouyang2022training,cao2023instruction}. Initially, this involved collecting and cleaning existing data, such as open-source NLP datasets~\cite{wang2023far,ding2023enhancing}. With the importance of instruction quality recognized, manual annotation methods emerged~\cite{wang2023far,zhou2024lima}. As larger datasets became necessary, approaches like Self-Instruct~\cite{wang2022self} used models to generate high-quality instructions~\cite{guo2024human}. However, complex instructions are rare, leading to strategies for synthesizing them by extending simpler ones~\cite{xu2023wizardlm,sun2024conifer,he2024can}. However, existing methods struggle with scalability and diversity.

\subsection{Back Translation}

Back-translation, a process of translating text from the target language back into the source language, is mainly used for data augmentation in tasks like machine translation~\cite{sennrich2015improving, hoang2018iterative}. ~\citet{li2023self} first applied this to large-scale instruction generation using unlabeled data, with Suri~\cite{pham2024suri} and Kun~\cite{zheng2024kun} extending it to long-form and Chinese instructions, respectively. ~\citet{nguyen2024better} enhanced this method by adding quality assessment to filter and revise data. Building on this, we further investigated methods to generate high-quality complex instruction dataset using back-translation.

\section{Approach}

\begin{figure*}[!t]
    \centering 
    \includegraphics[width=1.0\linewidth]{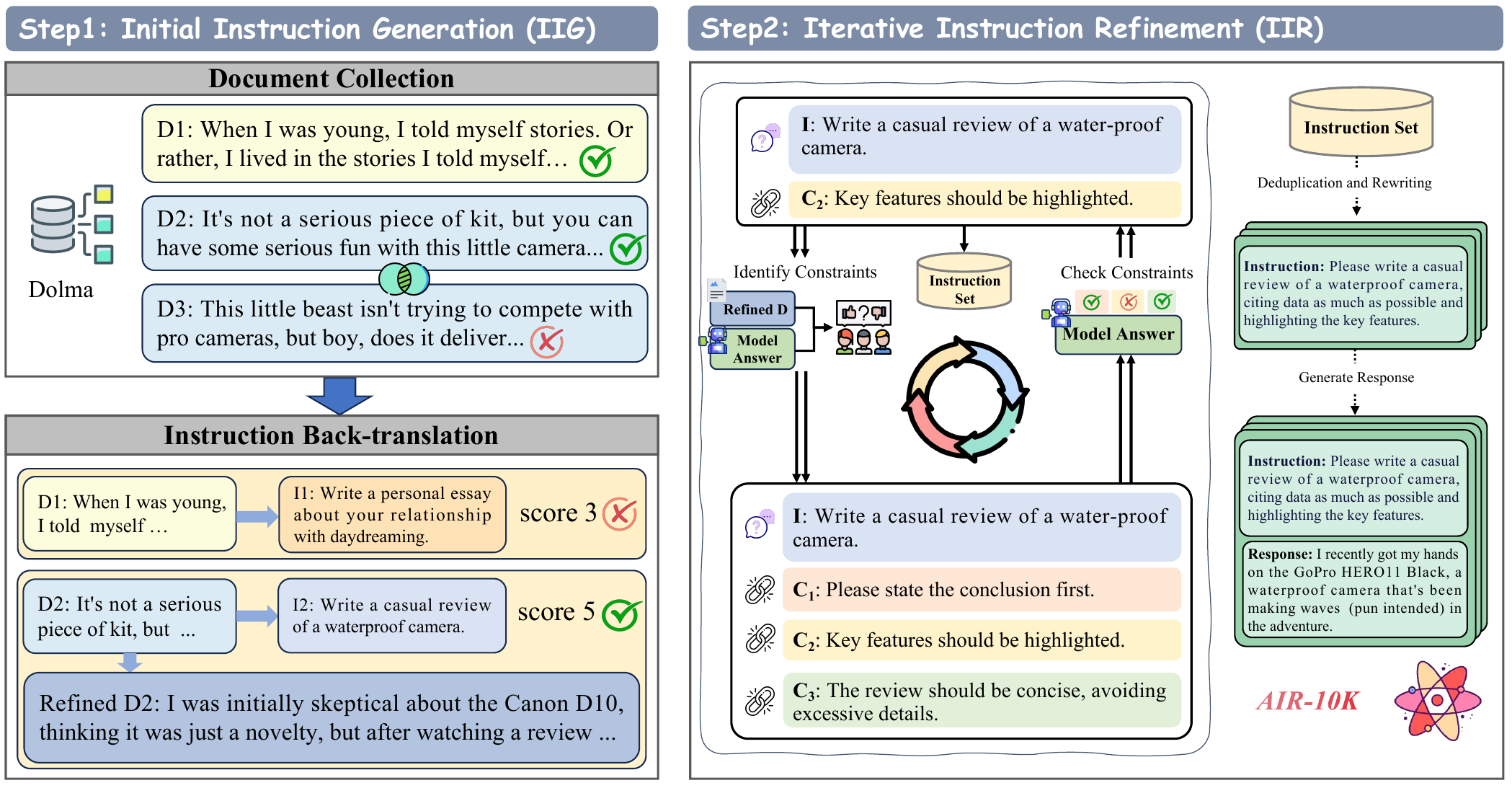}
    \vspace{-3mm}
    \caption{
     \textbf{AIR}: Automatic Iterative Refinement Framework. 
    }
    \vspace{-5mm}
    \label{fig:main}
\end{figure*}

Our approach mainly consists of two steps: 1) Initial Instruction Generation; 2) Iterative Instruction Refinement, as shown in Figure \ref{fig:main}. In this section, we will introduce the two steps in detail.

\subsection{Initial Instruction Generation (IIG)}

\paragraph{Document Collection.}
Traditional instruction generation methods such as Self-Instruct \cite{wang2022self} often suffer from limited diversity, as the generated instructions are generally re-combinations of the provided few-shot examples. Inspired by the work by ~\citet{li2023self}, we generate initial instructions using back translation based on human-written documents.


To further enhance the diversity of the generated instructions, we implement a density-based sampling mechanism for documents, as shown in Algorithm \ref{Alg:density}. Specifically, we convert documents into vector representations based on Sentence-Transformers\footnote{\texttt{sentence-transformers/all-MiniLM-L6-v2}.}, and perform sampling to maximize the density of samples in the representation space.

\begin{algorithm}[t]
  \renewcommand{\algorithmicrequire}{\textbf{Input:}}
  \renewcommand{\algorithmicensure}{\textbf{Output:}}
  \caption{Density-based Sampling}
  \label{alg:1}
  \begin{algorithmic}[1]
      \REQUIRE Instruction Dataset $D$ with $m$ samples.
      \ENSURE Selected Dataset $D'$ with $n$ samples.
      \STATE Derive the embeddings for each sample in $D$.
      \STATE Random sample one data point $x$ from $D$.
      \STATE Delete $x$ from $D$, add $x$ to $D'$.
      \FOR{$i = 1, 2, ..., t$}
          \STATE Calculate the cosine similarity score between $x$ and each sample from $D$.
          \STATE Select the least similar sample $x'$ from $D'$.
          \STATE Let $x$ = $x'$.
          \STATE Delete $x$ from $D$, add $x$ to $D'$. 
      \ENDFOR
  \end{algorithmic}
\label{Alg:density}
\end{algorithm}

In this way, we effectively eliminate redundant documents, enhancing the diversity of instructions. Moreover, this approach ensures that the knowledge introduced during instruction fine-tuning is evenly distributed across various domains. This not only prevents the model from overfitting to a specific domain but also mitigates the risk of catastrophic forgetting of fundamental capabilities.

Moreover, to further ensure the quality of the document collection, we filter out documents based on the following criteria: 1) Length: Documents with fewer than 50 words or exceeding 2,048 words are removed. 2) Symbol-to-text ratio: Documents where the proportion of symbols exceeds that of textual content are excluded. 3) Redundancy: Documents containing repetitive paragraphs or excessive symbol repetitions are eliminated.

\paragraph{Instruction Back-translation}
Based on the sampled documents, we employ the back-translation method to construct initial instructions.
Specifically, we utilize a guidance model to predict an instruction which can be accurately answered by (a portion of) the document\footnote{Detailed prompt templates are presented in Appendix \ref{appendix:prompt_iig}.}. This enables the generation of new instructions without relying on few-shot examples or pre-designed rules. Moreover, we can further ensure the diversity of the generated instructions by diversifying the documents.

However, despite being constructed from the document, the instruction do not always align well with the document in two key aspects \cite{nguyen2024better}. First, the document is unstructured and does not follow the AI-assistant format. Second, it may contain content irrelevant to the instruction. Therefore, we introduce an additional refinement step to transform the document into response format and remove irrelevant content.

To further ensure the quality of the instructions, we introduce a scoring step to filter out low-quality data. Each instruction is assigned a score on a scale of 1 to 5 by the guidance model, with each point corresponding to a specific aspect. Only instructions with a score greater than (or equal to) 4 are retained for the next step\footnote{Results of instruction score are presented in Appendix \ref{appendix:ins_score_case}.}.

\subsection{Iterative Instruction Refinement (IIR)}
\label{sec:IIR}
To enhance a model’s ability to follow complex instructions, it is crucial to construct complex instruction data that incorporates multiple constraints. Previous methods typically start with simple instructions and generate complex ones through rewriting or recombination~\cite{xu2023wizardlm}. However, the constraints generated in this way often do not meet actual needs or lack diversity.

An effective sample for complex instruction fine-tuning should adhere to two key principles:

\begin{enumerate}[itemsep=1mm, parsep=0pt]
    \item Whether the model’s response originally misaligns with constraint before it is added;
    \item Whether the model’s response still misaligns with the constraint after it is added.
\end{enumerate}

These constraints highlight the model’s weaknesses in handling complex instructions and require further improvement. Conversely, if a constraint does not meet these principles, it indicates that the constraint falls within the model’s current capabilities and does not require additional learning.

Therefore, we introduce constraint generation with LLM-as-judge guidance \cite{zheng2023judging}, which mimics the human process of iteratively refining prompts to form complex instructions\footnote{Detailed prompt templates are presented in Appendix \ref{appendix:prompt_iir}.}. As shown in Algorithm \ref{alg:iir}, during the process of iteration, we obtain the constraints that the model fails to satisfy, which require further fine-tuning.


\begin{algorithm}[h]
  \renewcommand{\algorithmicrequire}{\textbf{Input:}}
  \renewcommand{\algorithmicensure}{\textbf{Output:}}
  \caption{Iterative Instruction Refinement}
  \label{alg:1}
  \begin{algorithmic}[1]
      \REQUIRE Guidance model $M$, current model $m$, refined document $R$, initial instruction $I_0$.
      \ENSURE Constraint Sets $C_n$ and $C_n'$.
      \FOR{$i = 1, 2, ..., n$}
          \STATE Use $m$ to generate a response $A_i$ for the previous instruction $I_{i-1}$.
          \STATE Leverage $M$ as the judge, compare $A_i$ with $R$ to identify a new constraint $c_i$.
          \STATE Add $c_i$ to $C_n$.
          \STATE Add $c_i$ to $I_{i-1}$ to form a new instruction $I_i$.
          \STATE Use $m$ to generate a response $A_{i}'$ for $I_i$.
          \STATE Leverage $M$ as the judge, check whether $A_i'$ satisfies constraint $c_i$. If not, add $c_i$ to $C_n'$. 
      \ENDFOR
  \end{algorithmic}
\label{alg:iir}
\end{algorithm}


Throughout this process, as the number of constraints increases, the model's response also improves, making the identification of new constraints more challenging. To uncover constraints that better reflect human preferences, we use the refined document as the reference answer for the judgment process. Human-written documents inherently contain vast amounts of knowledge and formatting conventions that reflect human preferences. Therefore, the derived constraints will also align more closely with human preferences.

Finally, the constraint set is merged into a new complex instruction. Notice two constraint sets are derived: the first set $C_n$ satisfies Principle 1, while the second set $C_n'$, which includes an additional checking step, satisfies both Principle 1 and 2\footnote{The effect of the checking step is shown in Section \ref{sec:judge}.}.

While we leverage the refined document as the reference for the judgment process, it should not be used as the target for fine-tuning as in \citet{nguyen2024better}, as the document is not refined with the constraints presented explicitly. Therefore, we leverage the guidance model to re-generate the response based on the combined instruction\footnote{A detailed example illustrating the complete pipeline is provided in Appendix \ref{appendix:pipeline_case}.}.

\subsection{Data Statistics of AIR-10K}

\begin{figure}[t]
\centering
\subfigure[Distribution of domains]{
    \includegraphics[width=0.6\linewidth]{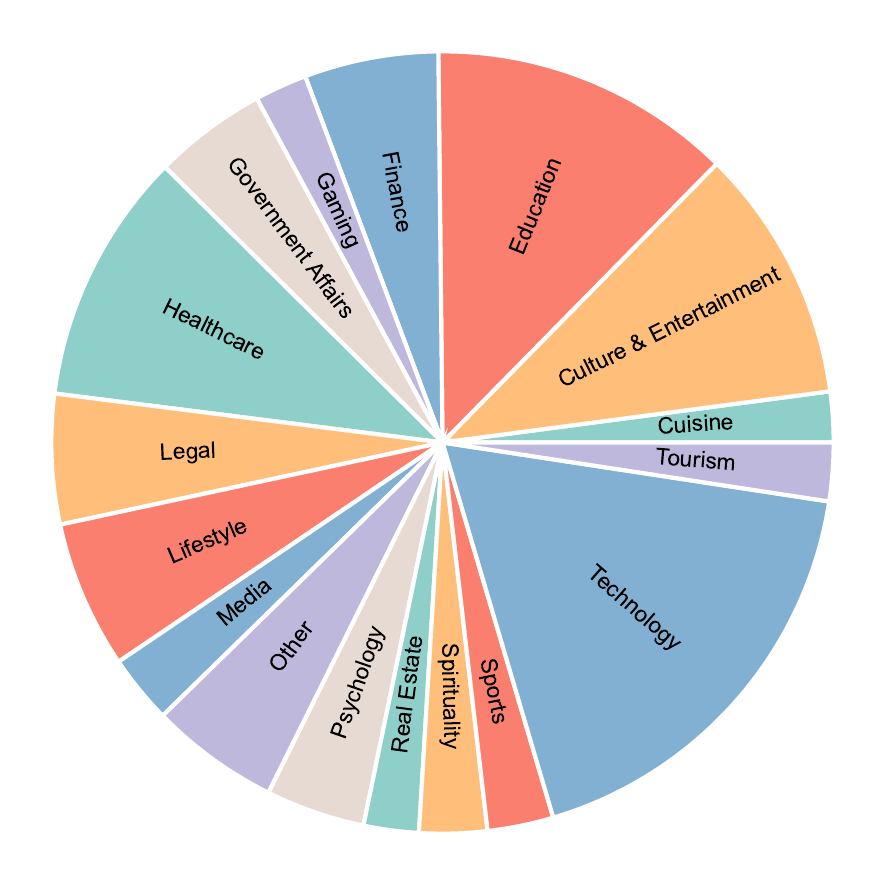}
}
\subfigure[Distribution of constraint types in iteration 1 and 5]{
    \includegraphics[width=0.95\linewidth]{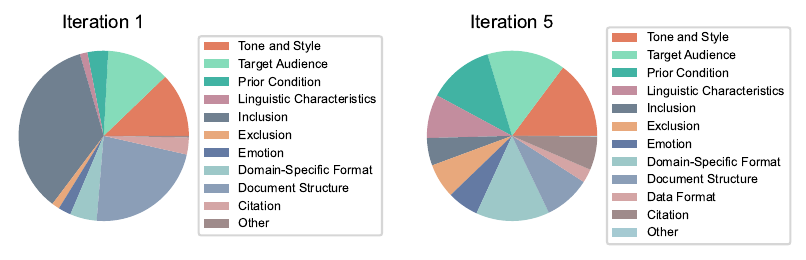}
}
\vspace{-2mm}
\caption{Data statistics of AIR-10K. }
\vspace{-5mm}
\label{figure:constraint_and_domain}
\end{figure}

\begin{figure}[t]
    \centering
    \includegraphics[width=0.4\textwidth]{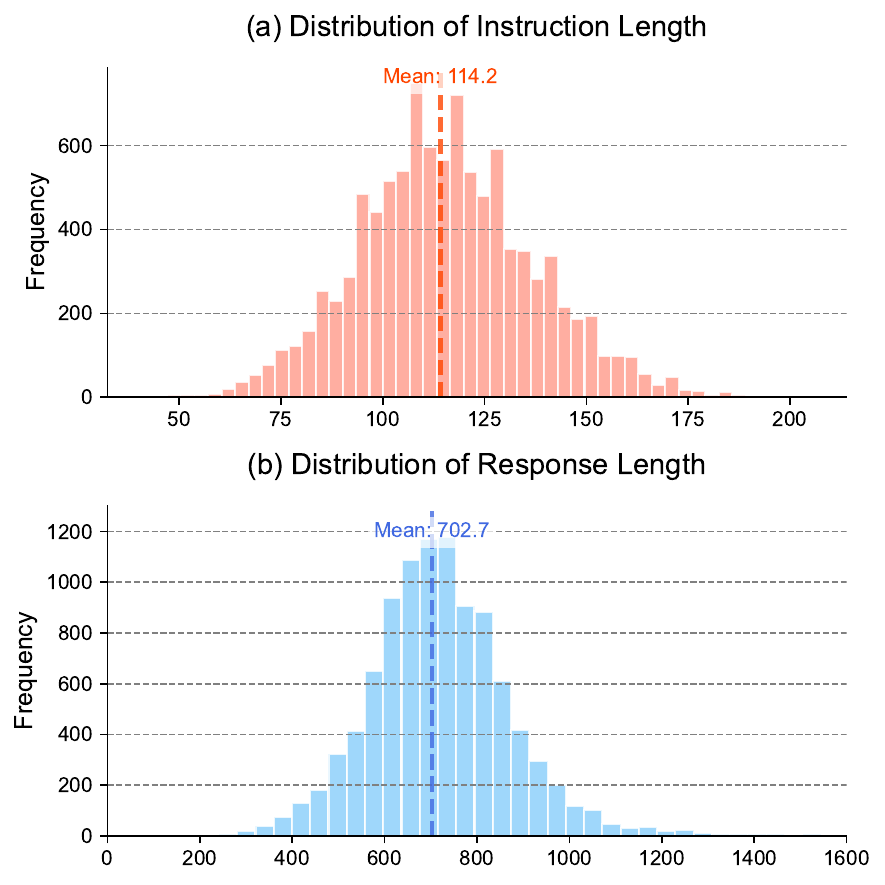}
    \vspace{-3mm}
    \caption{Length distribution of AIR-10K.}
    \vspace{-3mm}
    \label{figure:length_distribution}
\end{figure}

With our proposed framework, we constructed a high-quality complex instruction dataset, \textbf{AIR-10K}, based on openly available documents. We present the real-life scenario-specific domain distribution of AIR-10K in Figure \ref{figure:constraint_and_domain}(a). As can be seen, our dataset encompasses nearly 20 different domains in total, demonstrating a high degree of balance across diverse fields. Furthermore, we present the distribution of constraint types during iteration 1 and 5 in Figure \ref{figure:constraint_and_domain}(b). It is evident that in iteration 1, \textit{Inclusion} and \textit{Document Structure} constraints dominate. However, after four rounds of constraint additions, by iteration 5, the proportions of each constraint type become more uniform\footnote{The constraint type definition and complete distributions across all iterations are detailed in Appendix~\ref{appendix:constraints_type}.}.

We also analyze the length distributions of both instructions and responses. As shown in Figure \ref{figure:length_distribution}(a) and \ref{figure:length_distribution}(b), our instructions are of substantial information for capturing complex tasks.

\section{Experiments}
\subsection{Set-up}

\paragraph{Data.} 
Following \citet{nguyen2024better}, we utilize a subset of Dolma v1.7 \cite{dolma-2024} as the document source, which is derived from a collection of web pages and has undergone rigorous quality and content filtering to ensure data quality.

\paragraph{Models.} 
We apply our method on two models, Llama-3-8B and Qwen2.5-7B, and we apply preliminary supervised fine-tuning for both models. The preliminary fine-tuning process is conducted on two general instruction datas, namely ultrachat-200k \cite{ding2023enhancing} and tulu-330k \cite{lambert2024tulu3}, respectively. For the guidance model to construct the data, we rely on a larger model with the same group to ensure data quality, namely Qwen-2.5-72B-Instruct for Qwen-2.5-7B, and Llama-3-70B-Instruct for Llama-3-8B. We set the maximum number of iterations to 5.

\paragraph{Evaluation.} 
We mainly conduct evaluation on two complex instruction-following benchmarks, \textbf{CFBench}~\cite{zhang2024cfbench} and \textbf{FollowBench}~\cite{jiang2023followbench}, where instructions consist of multiple constraints. We also conduct evaluations on a general instruction benchmark of \textbf{AlpacaEval2}~\cite{dubois2024length}. Note that all benchmarks require GPT-4 for judgment, and we use GPT-4o-0806 \footnote{\url{platform.openai.com/docs/models/gp\#gpt-4o}} as the evaluator for all of them. We also conduct evaluation on fundamental capability benchmarks, including math, code, and knowledge tasks, and the results are presented in Appendix \ref{app: fundamental} due to space limitation.

\paragraph{Baselines.}
We mainly compare our method with four groups of methods as follows:

\begin{enumerate}[itemsep=1mm, parsep=0pt, leftmargin=*]
    \item \textbf{Human crafted instruction data}: This includes ShareGPT\footnote{\url{huggingface.co/datasets/anon8231489123/ShareGPT_Vicuna_unfiltered}}, which is a collections of real human-AI conversations.
    \item \textbf{Automatic crafted general instruction data}: This includes Self-Instruct \cite{wang2022self}, which leverages few-shot examples to self-generate simple instruction samples.
    \item \textbf{Automatic rewritten complex instruction data}: This includes Evol-Instruct \cite{xu2023wizardlm}, ISHEEP \cite{isheep}, Muffin \cite{lou2023muffin} and Conifer \cite{sun2024conifer}, which initiate with simple instructions and progressively construct more complex ones through rewriting or recombination.
    \item \textbf{Automatic back-translated complex instruction data}: This includes Suri \cite{pham2024suri} and Crab \cite{qi2024constraint}, which curate the complex instructions and constraints by back-translating the pre-existing response. These methods are the most closest to our work.
    
\end{enumerate}

Additionally, we also compare with the original back-translation \cite{cao2023instruction} and back-and-forth \cite{nguyen2024better}, where IIR is skipped and initial instructions are directly used.

Note that for all constructed datasets, we sample 10k instruction-response pairs for supervised fine-tuning under the same hyper-parameters\footnote{Detailed hyper-parameters are presented in Appendix \ref{appendix:hyper-parameters}.}.

\begin{table*}[!ht]
\centering
\setlength{\tabcolsep}{12pt}
\renewcommand{\arraystretch}{0.85}
\resizebox{0.98\textwidth}{!}{
\begin{tabular}{c|ccc|cc|cc}
\toprule
\multicolumn{8}{c}{\multirow{2}{*}{\raisebox{1.4em}{\makecell{\textbf{Fine-tuned on Llama-3-8B-UltraChat}}}}} \\
\midrule
\multirow{2}{*}{\raisebox{-0.4em}{\textbf{Method}}} & \multicolumn{3}{c}{\textbf{CF-Bench}}      & \multicolumn{2}{c}{\textbf{FollowBench}} & \multicolumn{2}{c}{\textbf{AlpacaEval2}} \\
\cmidrule(lr){2-4} \cmidrule(lr){5-6} \cmidrule(lr){7-8}
& \textbf{CSR} & \textbf{ISR} & \textbf{PSR} & \textbf{HSR}        & \textbf{SSR}       & \textbf{LC.}        & \textbf{Len}        \\
\midrule
Baseline    & 0.51         & 0.15         & 0.22         & 41.04               & 57.39              & 8.86               & 1,017     \\ 
\midrule
back-translation & $\text{0.40}_{\textcolor{deepred}{\text{-0.11}}}$ & $\text{0.11}_{\textcolor{deepred}{\text{-0.04}}}$ & $\text{0.15}_{\textcolor{deepred}{\text{-0.07}}}$ & $\text{21.19}_{\textcolor{deepred}{\text{-19.85}}}$ & $\text{33.92}_{\textcolor{deepred}{\text{-23.47}}}$ & $\text{0.96}_{\textcolor{deepred}{\text{-7.90}}}$ &  2,966    \\

back-and-forth & $\text{0.58}_{\textcolor{deepgreen}{\text{+0.07}}}$ & $\text{0.20}_{\textcolor{deepgreen}{\text{+0.05}}}$ & $\text{0.27}_{\textcolor{deepgreen}{\text{+0.05}}}$ & $\text{44.65}_{\textcolor{deepgreen}{\text{+3.61}}}$ & $\text{61.58}_{\textcolor{deepgreen}{\text{+4.19}}}$ & $\text{10.06}_{\textcolor{deepgreen}{\text{+1.20}}}$ &  1,440    \\
\midrule

ShareGPT & $\text{0.62}_{\textcolor{deepgreen}{\text{+0.11}}}$ & $\text{0.22}_{\textcolor{deepgreen}{\text{+0.07}}}$ & $\text{0.32}_{\textcolor{deepgreen}{\text{+0.10}}}$ & $\text{40.99}_{\textcolor{deepred}{\text{-0.05}}}$ & $\text{58.59}_{\textcolor{deepgreen}{\text{+1.20}}}$ & $\text{8.36}_{\textcolor{deepred}{\text{-0.50}}}$ &  1,052    \\
\midrule

Self-Instruct & $\text{0.34}_{\textcolor{deepred}{\text{-0.17}}}$ & $\text{0.08}_{\textcolor{deepred}{\text{-0.07}}}$ & $\text{0.10}_{\textcolor{deepred}{\text{-0.12}}}$ & $\text{12.33}_{\textcolor{deepred}{\text{-28.71}}}$ & $\text{26.92}_{\textcolor{deepred}{\text{-30.47}}}$ & $\text{2.76}_{\textcolor{deepred}{\text{-6.10}}}$ &  384     \\
\midrule

Evol-Instruct & $\text{0.57}_{\textcolor{deepgreen}{\text{+0.06}}}$ & $\text{0.22}_{\textcolor{deepgreen}{\text{+0.07}}}$ & $\text{0.28}_{\textcolor{deepgreen}{\text{+0.06}}}$ & $\text{43.58}_{\textcolor{deepgreen}{\text{+2.54}}}$ & $\text{59.21}_{\textcolor{deepgreen}{\text{+1.82}}}$ & $\text{7.15}_{\textcolor{deepred}{\text{-1.71}}}$ &  903         \\
MUFFIN & $\text{0.50}_{\textcolor{deepred}{\text{-0.01}}}$ & $\text{0.16}_{\textcolor{deepgreen}{\text{+0.01}}}$ & $\text{0.22}_{\textcolor{deepgreen}{\text{+0.00}}}$ & $\text{30.88}_{\textcolor{deepred}{\text{-10.16}}}$ & $\text{48.48}_{\textcolor{deepred}{\text{-8.91}}}$ & $\text{4.51}_{\textcolor{deepred}{\text{-4.35}}}$ &  791                  \\
Conifer & $\text{0.57}_{\textcolor{deepgreen}{\text{+0.06}}}$ & $\text{0.22}_{\textcolor{deepgreen}{\text{+0.07}}}$ & $\text{0.28}_{\textcolor{deepgreen}{\text{+0.06}}}$ & $\text{47.06}_{\textcolor{deepgreen}{\text{+6.02}}}$ & $\text{61.32}_{\textcolor{deepgreen}{\text{+3.93}}}$ & $\text{12.81}_{\textcolor{deepgreen}{\text{+3.95}}}$ &  1,084                 \\
I-SHEEP & $\text{0.53}_{\textcolor{deepgreen}{\text{+0.02}}}$ & $\text{0.17}_{\textcolor{deepgreen}{\text{+0.02}}}$ & $\text{0.23}_{\textcolor{deepgreen}{\text{+0.01}}}$ & $\text{34.26}_{\textcolor{deepred}{\text{-6.78}}}$ & $\text{50.28}_{\textcolor{deepred}{\text{-7.11}}}$ & $\text{5.41}_{\textcolor{deepred}{\text{-3.45}}}$ &  838                  \\
\midrule

Suri & $\text{0.26}_{\textcolor{deepred}{\text{-0.25}}}$ & $\text{0.05}_{\textcolor{deepred}{\text{-0.10}}}$ & $\text{0.07}_{\textcolor{deepred}{\text{-0.15}}}$ & $\text{3.19}_{\textcolor{deepred}{\text{-37.85}}}$ & $\text{3.83}_{\textcolor{deepred}{\text{-53.56}}}$ & $\text{0.60}_{\textcolor{deepred}{\text{-8.26}}}$ &  29         \\
Crab & $\text{0.56}_{\textcolor{deepgreen}{\text{+0.05}}}$ & $\text{0.18}_{\textcolor{deepgreen}{\text{+0.03}}}$ & $\text{0.25}_{\textcolor{deepgreen}{\text{+0.03}}}$ & $\text{39.92}_{\textcolor{deepred}{\text{-1.12}}}$ & $\text{56.83}_{\textcolor{deepred}{\text{-0.56}}}$ & $\text{9.05}_{\textcolor{deepgreen}{\text{+0.19}}}$ &  1,192                 \\

\midrule
\textbf{AIR} & \textbf{$\text{0.61}_{\textcolor{deepgreen}{\text{+0.10}}}$} & \textbf{$\text{0.24}_{\textcolor{deepgreen}{\text{+0.09}}}$} & \textbf{$\text{0.31}_{\textcolor{deepgreen}{\text{+0.09}}}$} & \textbf{$\text{50.69}_{\textcolor{deepgreen}{\text{+9.65}}}$} & \textbf{$\text{63.89}_{\textcolor{deepgreen}{\text{+6.50}}}$} & \textbf{$\text{21.00}_{\textcolor{deepgreen}{\text{+12.14}}}$} &  1,813        \\
\bottomrule
\multicolumn{8}{c}{\multirow{2}{*}{\raisebox{1.4em}{\makecell{\textbf{Fine-tuned on Qwen-2.5-7B-UltraChat}}}}} \\
\midrule
\multirow{2}{*}{\raisebox{-0.4em}{\textbf{Method}}} & \multicolumn{3}{c}{\textbf{CF-Bench}}      & \multicolumn{2}{c}{\textbf{FollowBench}} & \multicolumn{2}{c}{\textbf{AlpacaEval2}} \\
\cmidrule(lr){2-4} \cmidrule(lr){5-6} \cmidrule(lr){7-8}
& \textbf{CSR} & \textbf{ISR} & \textbf{PSR} & \textbf{HSR}        & \textbf{SSR}       & \textbf{LC.}        & \textbf{Len}        \\
\midrule
Baseline & 0.68 & 0.29 & 0.40 & 47.71 & 64.79 & 10.87 & 836 \\
\midrule
back-translation & $\text{0.42}_{\textcolor{deepred}{\text{-0.26}}}$ & $\text{0.14}_{\textcolor{deepred}{\text{-0.15}}}$ & $\text{0.18}_{\textcolor{deepred}{\text{-0.22}}}$ & $\text{21.62}_{\textcolor{deepred}{\text{-26.09}}}$ & $\text{34.86}_{\textcolor{deepred}{\text{-29.93}}}$ & $\text{1.79}_{\textcolor{deepred}{\text{-9.08}}}$ &  3,266  \\
back-and-forth & $\text{0.63}_{\textcolor{deepred}{\text{-0.05}}}$ & $\text{0.24}_{\textcolor{deepred}{\text{-0.05}}}$ & $\text{0.34}_{\textcolor{deepred}{\text{-0.06}}}$ & $\text{45.33}_{\textcolor{deepred}{\text{-2.38}}}$ & $\text{60.39}_{\textcolor{deepred}{\text{-4.40}}}$ & $\text{12.59}_{\textcolor{deepgreen}{\text{+1.72}}}$ &  1,480  \\
\midrule
ShareGPT & $\text{0.69}_{\textcolor{deepgreen}{\text{+0.01}}}$ & $\text{0.32}_{\textcolor{deepgreen}{\text{+0.03}}}$ & $\text{0.41}_{\textcolor{deepgreen}{\text{+0.01}}}$ & $\text{47.67}_{\textcolor{deepred}{\text{-0.04}}}$ & $\text{64.46}_{\textcolor{deepred}{\text{-0.33}}}$ & $\text{10.75}_{\textcolor{deepred}{\text{-0.12}}}$ &  1,028  \\
\midrule
Self-Instruct & $\text{0.39}_{\textcolor{deepred}{\text{-0.29}}}$ & $\text{0.10}_{\textcolor{deepred}{\text{-0.19}}}$ & $\text{0.14}_{\textcolor{deepred}{\text{-0.26}}}$ & $\text{20.10}_{\textcolor{deepred}{\text{-27.61}}}$ & $\text{35.47}_{\textcolor{deepred}{\text{-29.32}}}$ & $\text{2.47}_{\textcolor{deepred}{\text{-8.40}}}$ &  557  \\
\midrule
Evol-Instruct & $\text{0.67}_{\textcolor{deepred}{\text{-0.01}}}$ & $\text{0.30}_{\textcolor{deepgreen}{\text{+0.01}}}$ & $\text{0.40}_{\textcolor{deepgreen}{\text{+0.00}}}$ & $\text{46.67}_{\textcolor{deepred}{\text{-1.04}}}$ & $\text{63.98}_{\textcolor{deepred}{\text{-0.81}}}$ & $\text{8.81}_{\textcolor{deepred}{\text{-2.06}}}$ &  964  \\
MUFFIN & $\text{0.61}_{\textcolor{deepred}{\text{-0.07}}}$ & $\text{0.26}_{\textcolor{deepred}{\text{-0.03}}}$ & $\text{0.34}_{\textcolor{deepred}{\text{-0.06}}}$ & $\text{45.27}_{\textcolor{deepred}{\text{-2.44}}}$ & $\text{62.45}_{\textcolor{deepred}{\text{-2.34}}}$ & $\text{8.44}_{\textcolor{deepred}{\text{-2.43}}}$ &  880  \\
Conifer & $\text{0.70}_{\textcolor{deepgreen}{\text{+0.02}}}$ & $\text{0.34}_{\textcolor{deepgreen}{\text{+0.05}}}$ & $\text{0.44}_{\textcolor{deepgreen}{\text{+0.04}}}$ & $\text{51.65}_{\textcolor{deepgreen}{\text{+3.94}}}$ & $\text{65.72}_{\textcolor{deepgreen}{\text{+0.93}}}$ & $\text{19.39}_{\textcolor{deepgreen}{\text{+8.52}}}$ &  1,024  \\
I-SHEEP & $\text{0.63}_{\textcolor{deepred}{\text{-0.05}}}$ & $\text{0.25}_{\textcolor{deepred}{\text{-0.04}}}$ & $\text{0.36}_{\textcolor{deepred}{\text{-0.04}}}$ & $\text{41.96}_{\textcolor{deepred}{\text{-5.75}}}$ & $\text{59.48}_{\textcolor{deepred}{\text{-5.31}}}$ & $\text{6.43}_{\textcolor{deepred}{\text{-4.44}}}$ &  996  \\
\midrule
Suri & $\text{0.31}_{\textcolor{deepred}{\text{-0.37}}}$ & $\text{0.07}_{\textcolor{deepred}{\text{-0.22}}}$ & $\text{0.10}_{\textcolor{deepred}{\text{-0.30}}}$ & $\text{4.55}_{\textcolor{deepred}{\text{-43.16}}}$ & $\text{4.85}_{\textcolor{deepred}{\text{-59.94}}}$ & $\text{0.94}_{\textcolor{deepred}{\text{-9.93}}}$ &  239  \\
Crab & $\text{0.62}_{\textcolor{deepred}{\text{-0.06}}}$ & $\text{0.24}_{\textcolor{deepred}{\text{-0.05}}}$ & $\text{0.32}_{\textcolor{deepred}{\text{-0.08}}}$ & $\text{41.48}_{\textcolor{deepred}{\text{-6.23}}}$ & $\text{59.57}_{\textcolor{deepred}{\text{-5.22}}}$ & $\text{9.68}_{\textcolor{deepred}{\text{-1.19}}}$ &  1,102  \\
\midrule
\textbf{AIR} & \textbf{$\text{0.76}_{\textcolor{deepgreen}{\text{+0.08}}}$} & \textbf{$\text{0.41}_{\textcolor{deepgreen}{\text{+0.12}}}$} & \textbf{$\text{0.51}_{\textcolor{deepgreen}{\text{+0.11}}}$} & \textbf{$\text{59.07}_{\textcolor{deepgreen}{\text{+11.36}}}$} & \textbf{$\text{71.35}_{\textcolor{deepgreen}{\text{+6.56}}}$} & \textbf{$\text{32.43}_{\textcolor{deepgreen}{\text{+21.56}}}$} &  1,779  \\

\bottomrule
\end{tabular}}
\vspace{-3mm}
\caption{Experiment results on Llama-3-8B and Qwen-2.5-7B, with both models fine-tuned with ultrachat-200k \cite{ding2023enhancing}. Llama-3-70B-Instruct and Qwen-2.5-72B-Instruct are used as the guidance models respectively.}
\vspace{-3mm}
\label{table:main1}
\end{table*}

\begin{table}[!ht]
\centering
\renewcommand{\arraystretch}{0.95}
\resizebox{0.44\textwidth}{!}{
\begin{tabular}{c|ccc|cc}
\toprule
\multicolumn{6}{c}{\multirow{2}{*}{\raisebox{1.4em}{\makecell{\textbf{Fine-tuned on Llama-3-8B-Tulu}}}}} \\
\midrule
\multirow{2}{*}{\raisebox{-0.4em}{\textbf{Method}}}
& \multicolumn{3}{c|}{\textbf{CF-Bench}}   & \multicolumn{2}{c}{\textbf{AlpacaEval2}} \\ 
\cmidrule{2-6}
& \textbf{CSR} & \textbf{ISR} & \textbf{PSR}  & \textbf{LC.}  & \textbf{Len}  \\ 
\midrule
Baseline & 0.50 & 0.15 & 0.20 & 5.20 & 995 \\
\midrule
back-trans & 0.27 & 0.07 & 0.08 & 1.09 & 2,263 \\
back\&forth & 0.47 & 0.14 & 0.19 & 9.04 & 1,337 \\
\midrule
ShareGPT & 0.61 & 0.21 & 0.29 & 9.00 & 1,080 \\
\midrule
Self-Instruct & 0.30 & 0.07 & 0.09 & 2.63 & 378 \\
\midrule
Evol-Instruct & 0.58 & 0.19 & 0.27 & 18.09 & 991 \\
MUFFIN & 0.46 & 0.15 & 0.18 & 5.21 & 760 \\
Conifer & 0.61 & 0.24 & 0.32 & 7.15 & 903 \\
I-SHEEP & 0.49 & 0.16 & 0.19 & 3.11 & 931 \\
\midrule
Crab & 0.56 & 0.19 & 0.27 & 8.55 & 1,221 \\
\midrule
\textbf{AIR} & \textbf{0.68} & \textbf{0.28} & \textbf{0.38} & \textbf{22.00} & 2,097 \\
\bottomrule
\end{tabular}}
\vspace{-3mm}
\caption{Experiment results on Llama-3-8B, fine-tuned with tulu-330k \cite{lambert2024tulu3}, with Llama-3-70B-Instruct as the guidance model.}
\vspace{-3mm}
\label{table:main2}
\end{table}

\subsection{Main Results}
As shown in Tables \ref{table:main1} and \ref{table:main2}, our proposed method achieves the best performance on both complex and general instruction-following benchmarks, demonstrating its effectiveness. In contrast, automatically crafted general instruction data significantly underperform, highlighting the importance of multiple constraints in effective instruction fine-tuning. Automatic rewritten instructions also underperform, as their constructed constraints do not align with real-world practice. Additionally, automatically back-translated instructions underperform as well. Despite the constraints being derived from documents, the documents (even after refinement) suffer from misalignment and should not be direct used as the target for fine-tuning.


\subsection{Data Quality Evaluation}

\begin{figure}[t]
\centering
\includegraphics[width=0.9\linewidth]
{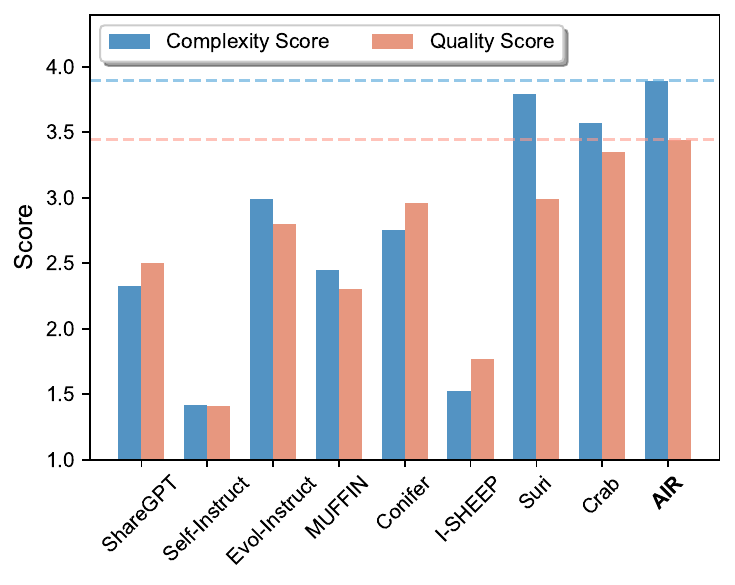}
\vspace{-3mm}
\caption{Comparison of averaged complexity and quality scores on different datasets.}
\vspace{-3mm}
\label{figure:complexity_scores}
\end{figure}

To evaluate our dataset's quality, we employed the Deita scorer \cite{liu2024Deita}, which utilizes LLM to assess complexity score for instructions and quality score for both instructions and responses. As shown in Figure \ref{figure:complexity_scores}, our approach significantly outperforms human crafted instructions, automatically crafted general instructions, and automatically rewritten complex instructions in terms of both complexity and quality scores. Notably, our method shows marginal improvements over automatic back-translation approaches like Suri and Crab, despite their use of high-quality seed datasets (e.g., Alpaca GPT4 for Crab) and advanced models (e.g., GPT-4-turbo for Suri). These results validate the effectiveness of our data generation strategy.

\begin{figure}[t]
\centering
\subfigure[Diversity: unique trigrams and token length]{
    \includegraphics[width=0.90\linewidth]{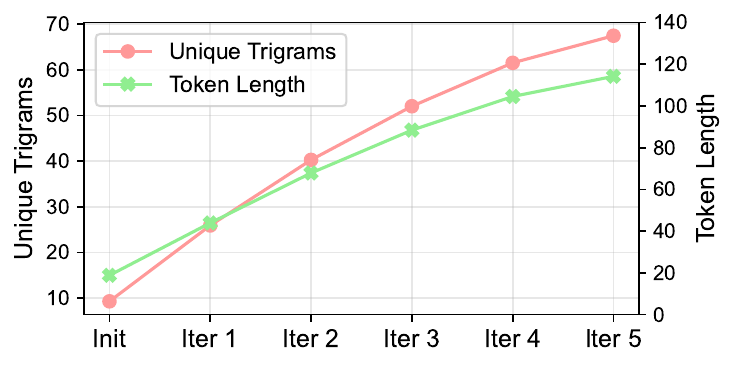}
}
\vspace{-3mm}
\subfigure[Complexity and quality score]{
    \includegraphics[width=0.90\linewidth]{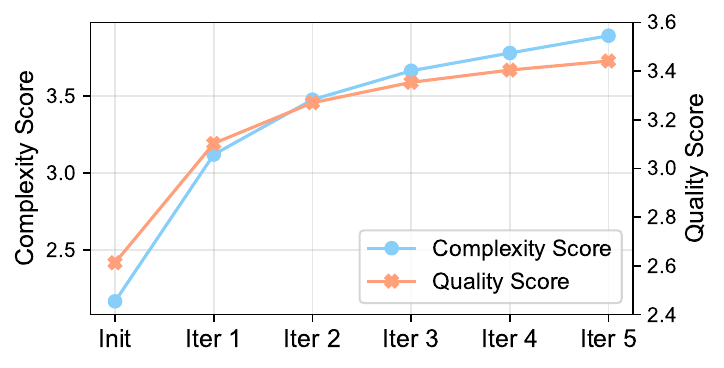}
}
\caption{Variation of quality indicators across iterations. \textit{Init} represents initial instructions generated through the IIG step.}
\vspace{-3mm}
\label{figure:instruction_and_complexity}
\end{figure}

To investigate the effect of iterative refinement, we analyze the variation of average unique trigrams and token lengths across iterations in Figure \ref{figure:instruction_and_complexity}(a). The results demonstrate consistent increases in both instruction length and unique trigrams, indicating that newly added constraints is diverse rather than mere repetition. Furthermore, Figure \ref{figure:instruction_and_complexity}(b) displays the evolution of complexity and quality scores throughout the iterations, showing steady improvement of data quality as the iterations progress.




\subsection{Judgment Strategy for Better Constraint}
\label{sec:judge}

In this section, we investigate the optimal judgment strategy for constraint generation. When humans adjust prompts based on the output, they typically have a pre-expected response as the reference in mind, and constraints are issued to guide the response closer to the reference. Therefore, we compare three judgment settings: 1) No judgment, directly curate constraints; 2) Judge without document as the reference. Instead, use the guidance models' response as the reference; 3) Judge with the refined document as the reference. 

As shown in Table \ref{table:judge}, the judgment process is essential for uncovering valuable constraints to improve the complex instruction following ability. LLM-judge can curate constraints that reflects the insufficiency of the model which requires further tuning. Moreover, using document as reference is also essential due to the limited judgment ability of the model, and human-written references aid in more targeted constraint construction.

On the other hand, the additional checking step does not improve complex instruction-following ability, as the checking step would result in fewer constraints. However, we observe improved performance on general-instruction following, indicating there exists a trade-off between general and complex instruction following abilities.

\begin{table}[!ht]
\centering
\renewcommand{\arraystretch}{0.95}
\resizebox{0.44\textwidth}{!}{
\begin{tabular}{c|cc|cc}
\toprule
\multirow{2}{*}{\textbf{Method}} & \multicolumn{2}{c|}{\textbf{FollowBench}} & \multicolumn{2}{c}{\textbf{AlpacaEval2}} \\
& \textbf{HSR} & \textbf{SSR} & \textbf{LC.} & \textbf{Len} \\
\midrule
\multicolumn{5}{l}{\textit{Results on Llama-3-8B-UltraChat}} \\
\midrule
Baseline & 41.04 & 57.39 & 8.86 & 1,017 \\
\midrule
w/o judge & 47.15 & 62.62 & 19.07 & 1,706 \\
judge w/o doc & 51.24 & 63.81 & 20.00 & 1,717 \\
judge w/ doc & \textbf{52.34} & \textbf{64.09} & 19.74 & 1,408 \\
\midrule
w/ check & 50.69 & 63.89 & \textbf{21.00} & 1,813 \\
\midrule
\multicolumn{5}{l}{\textit{Results on Llama-3-8B-Tulu}} \\
\midrule
Baseline & 34.91 & 51.76 & 5.20 & 995 \\
\midrule	
w/o judge & 47.59 & 63.60 & 18.32 & 2,067 \\
judge w/o doc & 50.62 & 63.69 & 17.02 & 2,842 \\
judge w/ doc & \textbf{54.16} & \textbf{67.52} & 20.45 & 1,639 \\
\midrule
w/ check & 51.35 & 66.09 & \textbf{21.09} & 2,049 \\
\bottomrule
\end{tabular}}
\caption{Experiment results on Llama-3-8B models with constraints from different judgment strategies.}
\label{table:judge}
\end{table}

\subsection{Influnce of Iterative Judge}

\begin{figure}[t]
\centering
\subfigure[Llama-3-8B-UltraChat]{
    \includegraphics[width=0.95\linewidth]{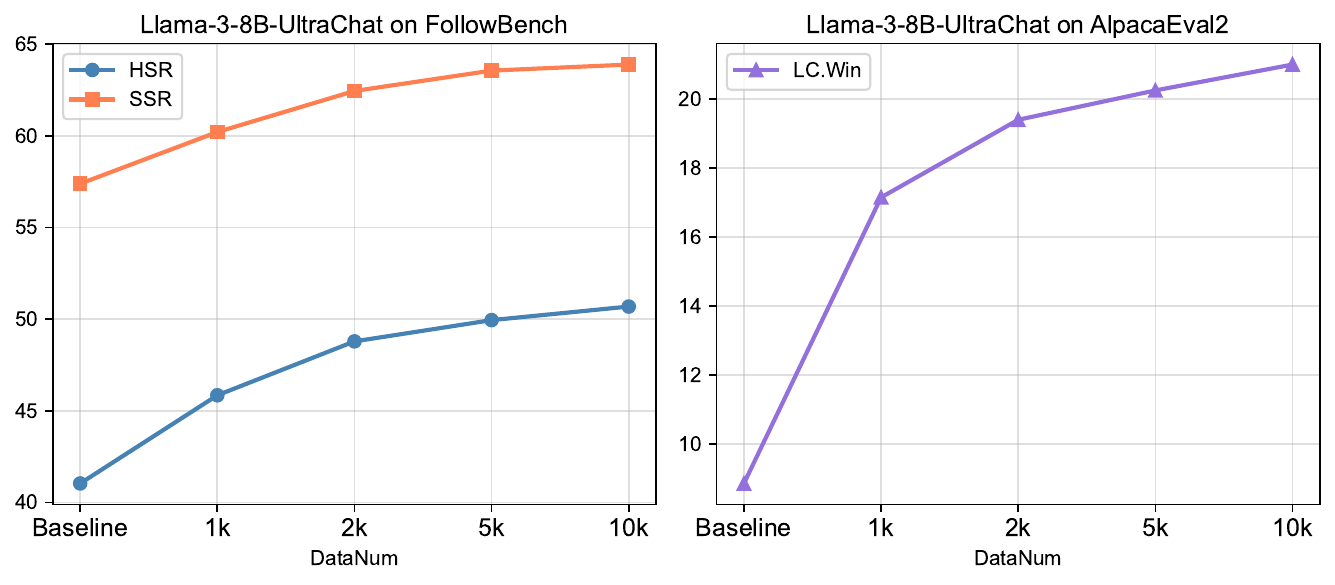}
}
\subfigure[Llama-3-8B-Tulu]{
    \includegraphics[width=0.95\linewidth]{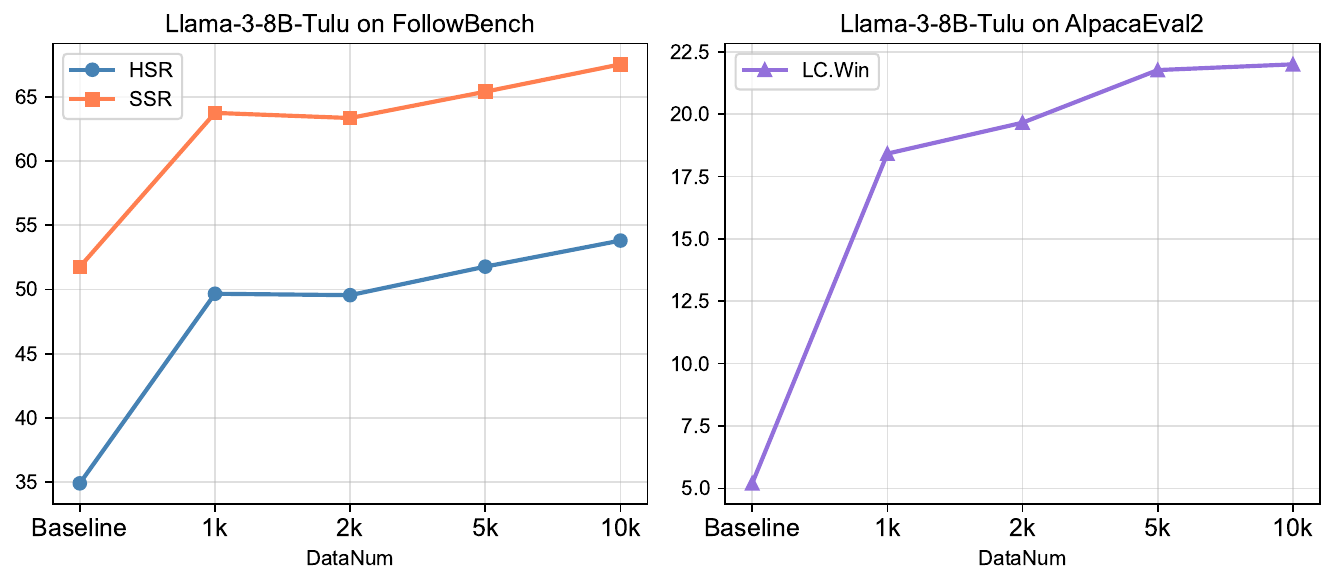}
}
\vspace{-3mm}
\caption{The variation of performance on FollowBench and AlpacaEval2 with the variation of data number.}
\vspace{-3mm}
\label{figure:data_num}
\end{figure}

\begin{table}[h]
\centering
\renewcommand{\arraystretch}{0.95}
\resizebox{0.4\textwidth}{!}{
\begin{tabular}{c|cc|cc}
\toprule
\multirow{2}{*}{\textbf{Iteration}} & \multicolumn{2}{c|}{\textbf{FollowBench}} & \multicolumn{2}{c}{\textbf{AlpacaEval2}} \\
& \textbf{HSR}        & \textbf{SSR}       & \textbf{LC.}      & \textbf{Len} \\ 
\midrule
\begin{tabular}[c]{@{}c@{}}Baseline\end{tabular}  & 34.91       & 51.76      & 5.20       & 995         \\
\midrule Init    & 46.37   & 61.87    & 17.96  & 1,602    \\
1    & 49.75   & 64.78    & 21.63  & 1,994    \\
2   & 53.82    & 67.55    & 21.01     & 1,829  \\
3   & \textbf{54.46}        & 67.54     & 20.69   & 1,722     \\
4   & 53.97       & 67.09    & \textbf{22.50}      & 1,672     \\
5   & 53.30     & \textbf{67.91}    & 20.78    & 1,599       \\
\bottomrule
\end{tabular}}
\vspace{-3mm}
\caption{Experiment results on Llama-3-8B-Tulu fine-tuned on different iterations. \textit{Init} represents initial instructions generated through the IIG step.}
\vspace{-3mm}
\label{table:judge_iterations}
\end{table}

In this section, we investigate the effectiveness of iterative judge by examining model performance across different iterations. As shown in Table \ref{table:judge_iterations}, the iterative judge process demonstrates clear benefits compared to both the baseline and IIG step.

Specifically, we observe consistent improvements on FollowBench and AlpacaEval2 through the first two iterations. This suggests that the iterative judging process effectively identifies and incorporates increasingly sophisticated constraints that are valuable for complex instruction following. However, improvements tend to plateau after the third iteration. This could be attributed to the fact that the most critical and fundamental constraints have already been discovered in earlier iterations.



\subsection{Influence of Data Quantity}

\indent In this section, we investigate the impact of data quantity on AIR's performance. We present the results of models trained with varying amounts of data in Figure \ref{figure:data_num}. As shown, performance on both general and complex instruction tasks improves with increasing data quantity. On the other hand, the model can achieve superior performance with only 1k training samples, and the performance gains become marginal as more data is added. Therefore, in practical applications, the optimal amount of fine-tuning data can be determined based on available computational resources.

\subsection{Influence of Guidance Model Size}

\begin{table}[!ht]
\centering
\renewcommand{\arraystretch}{0.95}
\resizebox{0.4\textwidth}{!}{
\begin{tabular}{c|cc|cc}
\toprule
\multirow{2}{*}{\textbf{Guid. Model}} & \multicolumn{2}{c}{\textbf{FollowBench}} & \multicolumn{2}{c}{\textbf{AlpacaEval2}} \\
& \textbf{HSR}   & \textbf{SSR}   & \textbf{LC.}      & \textbf{Len}  \\
\midrule
\begin{tabular}[c]{@{}c@{}}Baseline\end{tabular}  & 47.71    & 64.79  & 10.87   & 836   \\
\midrule
14B     & 57.72   & 70.59   & 29.13   & 1,501   \\
32B     & \textbf{60.06}   & \textbf{71.97}   & 26.39   & 1,309    \\
72B     & 59.07   & 71.35   & \textbf{32.43}  & 1,779     \\
\bottomrule
\end{tabular}}
\caption{Experiment results on Qwen-2.5-7B-UltraChat fine-tuned with different guidance model size.}
\label{table:guidance_size}
\end{table}

In Table \ref{table:guidance_size}, we investigate the impact of guidance model size on AIR's performance. We performed experiments with Qwen-2.5-7B-UltraChat as the base model, while varying the guidance model size from 14B to 72B parameters. As shown, all guidance models significantly improve instruction-following ability compared to the baseline, while larger models generally present more improvement. On the other hand, even the 14B guidance model demonstrates remarkable improvement. This scalability across different model sizes highlights the robustness and efficiency of our approach.


\section{Conclusion}
This paper introduces the Automatic Iterative Refinement (AIR) framework, a novel approach for generating complex instructions that better align with real-world scenarios. The framework employs an iterative refinement process guided by LLM-as-judge to generate high-quality complex constraints. We also construct a complex instruction dataset, AIR-10K, to facilitate the research and application of complex instruction following.

While previous methods for complex instruction following often introduce constraints without clear justification, it is crucial to understand what authentic complex instruction entails. In the future, we will conduct further research on the effectiveness and efficiency of complex instruction data.


\section*{Limitations}

Our work has several limitations. 1) Although our evaluation includes multiple established benchmarks and metrics, including human evaluation could further improve its credibility. Due to time and resource limitation, we have to leave this as future work. 2) Despite meticulous preprocessing, the Dolma dataset remains relatively noisy. Incorporating more high-quality documents (for example, judicial documents made public) could provide more knowledge and formality to support constraint construction. 3) The iterative nature of our framework requires multiple rounds of model inference, resulting in higher computational demands. While our ablation studies demonstrate effectiveness even with smaller guidance models and fewer samples, the computational cost remains a challenge for researchers with limited resources.

\section*{Ethical Considerations}
Our data construction framework primarily leverages proprietary models such as Llama-3-70B-Instruct, which have undergone extensive preference optimization to minimize the likelihood of generating instructions that raise ethical concerns. However, large-scale web corpora—our primary data sources—are uncensored and may contain harmful or toxic content. To address this, we recommend implementing more rigorous and meticulous filtering mechanisms to proactively identify and remove such instances if possible.

While the AIR framework mainly aims to enhance models' ability to follow complex instructions, it is important to note that some user constraints may conflict with system constraints set by developers. For example, users may request the generation of harmful or toxic content. Although our study does not specifically investigate conflicting constraints, there is a potential risk that the pipeline could prioritize user requests over developer-defined safeguards.

\bibliography{custom}
\clearpage
\appendix
\onecolumn

\section{Impact on Fundamental Capabilities}
\label{app: fundamental}

\begin{table}[h]
\centering
\renewcommand{\arraystretch}{1.0}
\resizebox{0.6\textwidth}{!}{
\begin{tabular}{c|ccccc|c}
\toprule
\textbf{Method} & \textbf{MMLU} & \textbf{CQA} & \textbf{NQ} & \textbf{HumanEval} & \textbf{GSM8K} & \textbf{AVG} \\
\midrule
\multicolumn{7}{l}{\textit{Results on Llama-3-8B-UltraChat}} \\
\midrule
Baseline & 64.00 & 72.97 & 29.61 & 30.49 & 57.47 & 50.90 \\
\midrule
AIR & 61.64 & \textbf{73.63} & \textbf{30.54} & 29.88 & 54.59 & 50.05 \\
\midrule
\multicolumn{7}{l}{\textit{Results on Qwen-2.5-7B-UltraChat}} \\
\midrule
Baseline & 73.64 & 82.39 & 25.68 & 52.20 & 81.65 & 63.11 \\
\midrule
AIR & 73.35 & \textbf{82.56} & \textbf{25.76} & \textbf{55.49} & \textbf{84.38} & \textbf{64.30} \\
\midrule
\multicolumn{7}{l}{\textit{Results on Llama-3-8B-Tulu}} \\
\midrule
Baseline & 65.43 & 79.44 & 32.22 & 50.61 & 64.14 & 58.36 \\
\midrule
AIR & 64.95 & \textbf{79.92} & \textbf{34.62} & \textbf{50.85} & 63.70 & \textbf{58.80} \\
\bottomrule
\end{tabular}}
\caption{Experiment results on fundamental capabilities.}
\label{table:fundamental}
\end{table}

Previous methods have shown LLMs may suffer from capability degradation during alignment~\cite{ouyang2022training}. To evaluate this concern, we tested our AIR method on MMLU \cite{hendrycks2021measuring}, CommonsenseQA (CQA) \cite{talmor-etal-2019-commonsenseqa}, Natural Questions (NQ) \cite{kwiatkowski2019natural}, HumanEval \cite{chen2021evaluatinglargelanguagemodels}, and GSM8K \cite{cobbe2021trainingverifierssolvemath}. In Table \ref{table:fundamental}, our method does not have a negative impact on fundamental capabilities. For Qwen-2.5-7B-UltraChat and Llama-3-8B-Tulu, our method even improves the average performance by 1.19 and 0.44 points, respectively. This indicates that instruction constructed from documents with evenly sampled distributions also present even distribution, which would not lead to catastrophic forgetting of fundamental capabilities.

\section{Case Study for Complete Pipeline }
\label{appendix:pipeline_case}

\begin{figure}[h]
\centering
\includegraphics[width=0.9\linewidth]{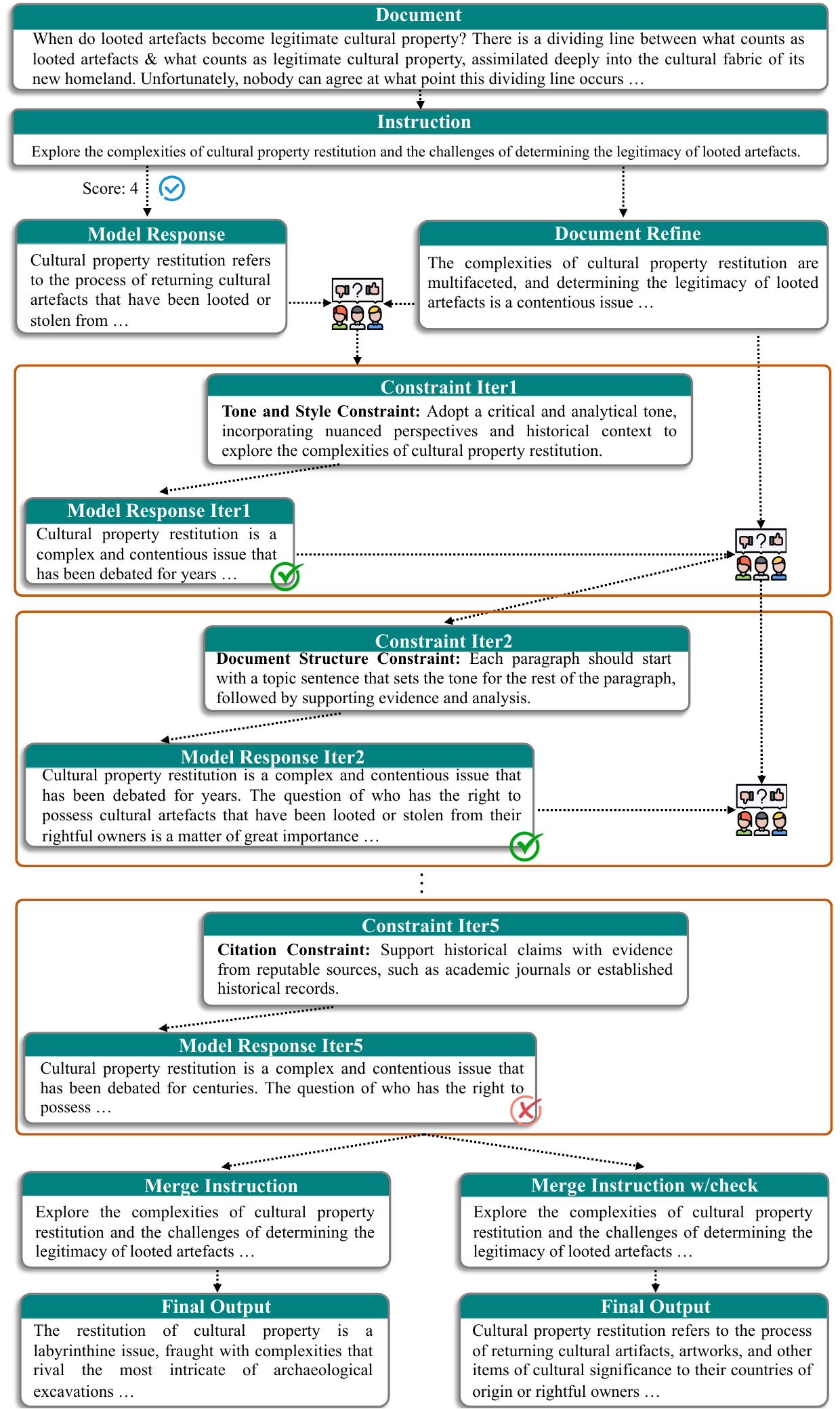}
\caption{End-to-End Pipeline Implementation Example.}
\label{figure:pipeline_case}
\end{figure}

This section presents a detailed end-to-end demonstration of our pipeline in Figure \ref{figure:pipeline_case}. The case study provides a thorough walkthrough of each stage in our instruction generation and refinement process.

\section{Constraint Type Taxonomy and Distribution Analysis}
\label{appendix:constraints_type}

\begin{figure}[h]
\centering
\includegraphics[width=0.9\linewidth]{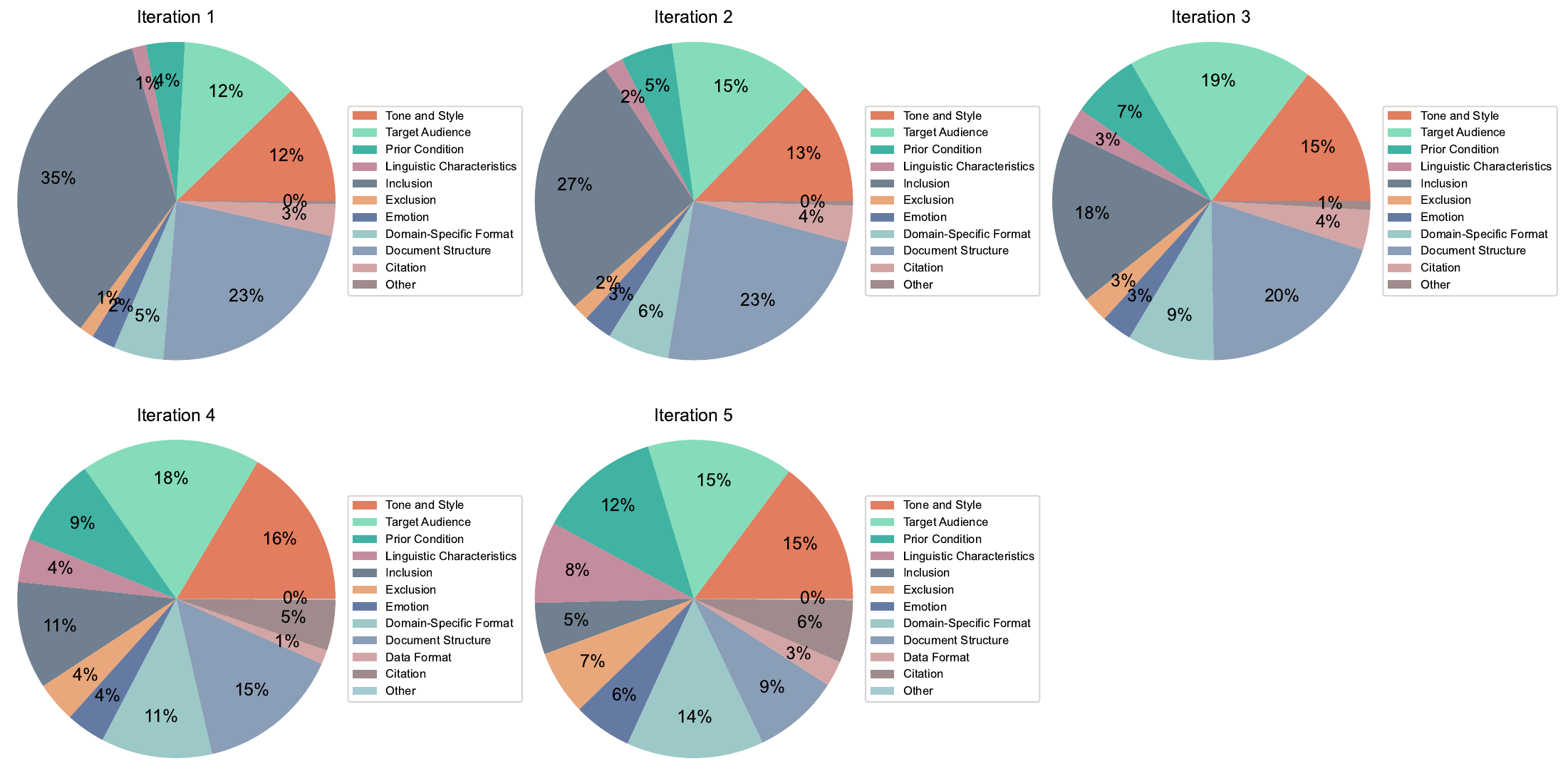}
\caption{Distribution of constraint types across all iterations.}
\label{figure:model_comparison}
\end{figure}

This section provides a detailed classification of constraint types, as defined in Table \ref{table:constraint_types}. Additionally, we present a comprehensive analysis of constraint type distribution patterns observed across five iterative refinement rounds, as visualized in Figure \ref{figure:model_comparison}.

\begin{table}[h]
\centering
\renewcommand{\arraystretch}{0.95}
\resizebox{0.85\textwidth}{!}{
\begin{tabular}{>{\raggedright\arraybackslash}m{0.25\linewidth}m{0.65\linewidth}}
\toprule
\textbf{Constraint Type} & \textbf{Description} \\
\midrule
Data Format & The generated content should conform to specific data structure formats, such as JSON, Markdown, Table, CSV, etc. \\
\midrule
Document Structure & The generated content should follow specific document organization patterns, including Numbered lists (1, 2, 3 or I, II, III), Bullet points (•, -, *), Custom templates with predefined sections, Tables, Headers, etc. \\
\midrule
Domain-Specific Format & Content must follow strict format rules for different industries \\
\midrule
Inclusion & Identify and list the specific elements or information that should be included in the generated content \\
\midrule
Exclusion & Identify and list the specific elements or information that should not be included in the generated content \\
\midrule
Citation & The generated content should include citations to sources, providing reliable sources and literature support; follow specific citation formats or reference styles \\
\midrule
Prior Condition & When a specific intention is met, a particular process should be followed to perform an operation or output specific content \\
\midrule
Target Audience & The generated content should target a specific audience, which affects the terminology used, the level of detail provided, and the complexity of the content \\
\midrule
Tone and Style & The generated content should adopt a specific tone and style, such as formal, polite, academic, concise, literary, romantic, or sci-fi \\
\midrule
Emotion & The generated content should express a specific emotion or mood, such as ensuring the content is positive, inspiring, or empathetic \\
\midrule
Linguistic Characteristics & Use specific linguistic features, such as metaphors, personification, and other rhetorical devices \\
\midrule
Multilingual & The generated content should be written in a specific language, such as English, Mandarin, or Spanish \\
\bottomrule
\end{tabular}}
\caption{Types of Constraints Used in Dataset Generation.}
\label{table:constraint_types}
\end{table}


\section{Model Training Hyper-parameters}
\label{appendix:hyper-parameters}

This section details our model training configuration based on the LlamaFactory \cite{zheng-etal-2024-llamafactory} framework. We employed Supervised Fine-Tuning (SFT) with hype-rparameters as outlined in Table \ref{table:hyperparams}.

\begin{table}[!h]
\centering
\resizebox{0.5\textwidth}{!}{
\begin{tabular}{l|cc}
\hline
\textbf{Configuration} & \textbf{Llama-3-8B}   & \textbf{Qwen-2.5-7B}   \\ \hline
max length             & 4096                  & 4096                   \\
learning rate          & 1e-5                  & 1e-5                   \\
scheduler              & cosine decay          & cosine decay           \\
training epochs        & 3                     & 3                      \\
batch size             & 64                    & 64                     \\
flash-attn             & fa2                   & fa2                  \\
numerical precision    & bf16                  & bf16                   \\
ZeRO optimizer         & stage 2               & stage 2                \\ \hline
\end{tabular}}
\caption{Hyper-parameters for Supervised Fine-Tuning.}
\label{table:hyperparams}
\end{table}

\section{Prompts for Initial Instruction Generation}
\label{appendix:prompt_iig}

\begin{figure}[h]
\centering
\includegraphics[width=0.8\linewidth]{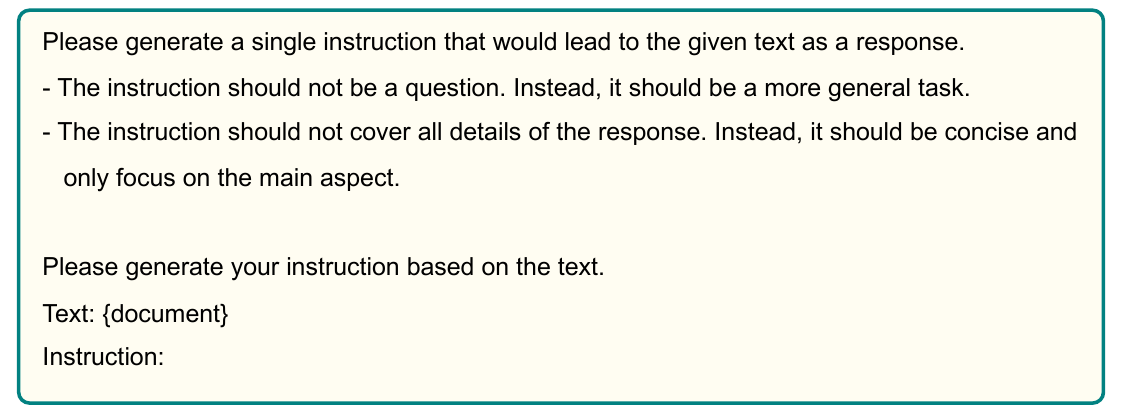}
\caption{Prompt for generating initial instructions through back-translation.}
\label{figure:prompt_back_ins}
\end{figure}

\begin{figure}[h]
\centering
\includegraphics[width=0.8\linewidth]{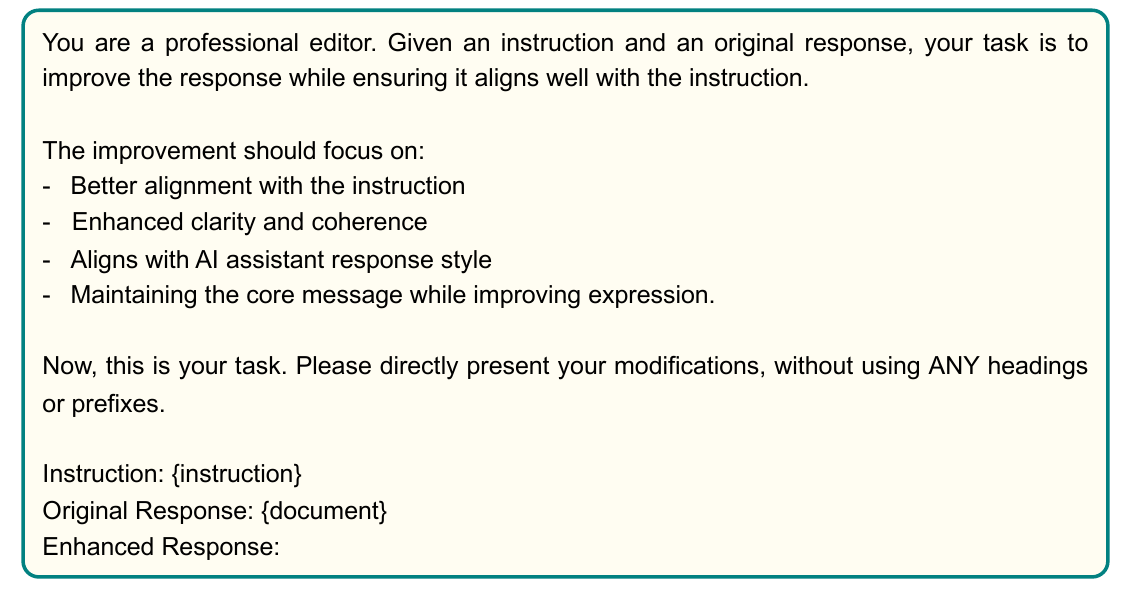}
\caption{Prompt for refining document content.}
\label{figure:prompt_document_polish}
\end{figure}

\begin{figure}[h]
\centering
\includegraphics[width=0.8\linewidth]{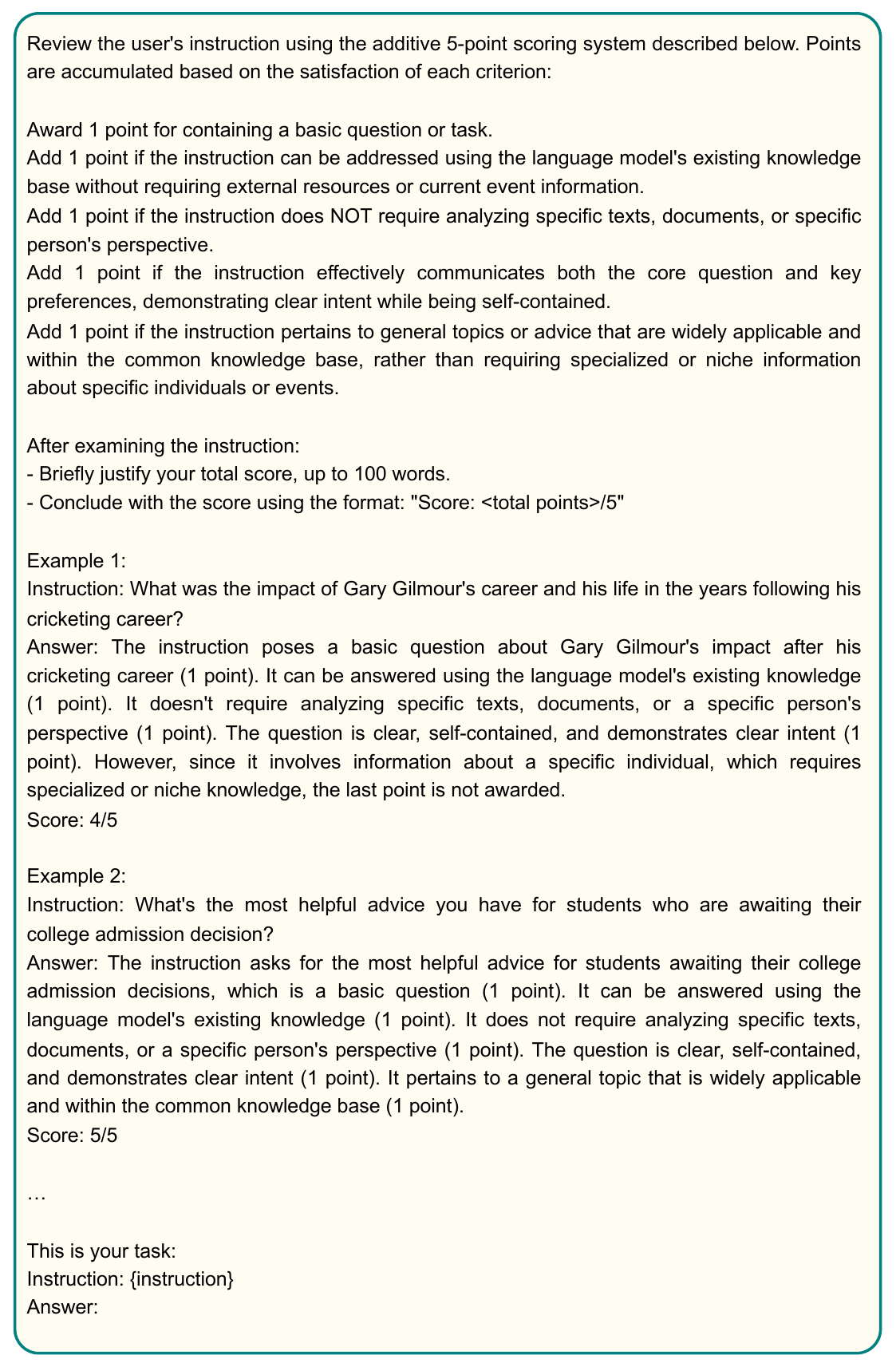}
\caption{Prompt for scoring initial instructions.}
\label{figure:prompt_ins_score}
\end{figure}

This section presents the prompts used in our data generation pipeline in Initial Instruction Generation step. These prompts serve different purposes in our methodology, from initial instruction generation through back-translation (Figure \ref{figure:prompt_back_ins}) to document refining (Figure \ref{figure:prompt_document_polish}) and instruction scoring (Figure \ref{figure:prompt_ins_score}). 

\section{Prompts for Iterative Instruction Refinement}
\label{appendix:prompt_iir}

\begin{figure}[h]
\centering
\includegraphics[width=0.8\linewidth]{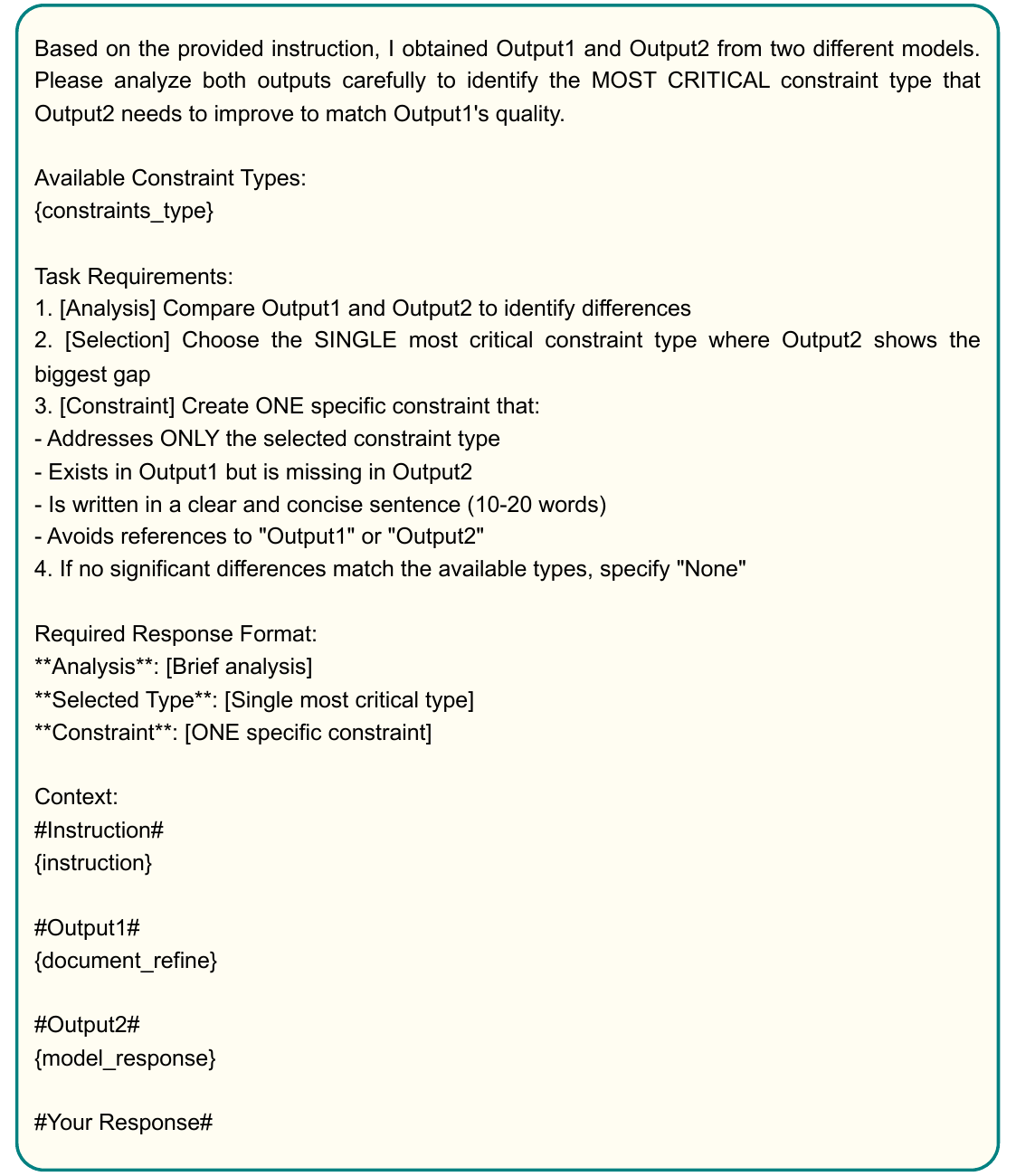}
\caption{Prompt for generating constraints based on judge.}
\label{figure:cons_generate}
\end{figure}

\begin{figure}[h]
\centering
\includegraphics[width=0.8\linewidth]{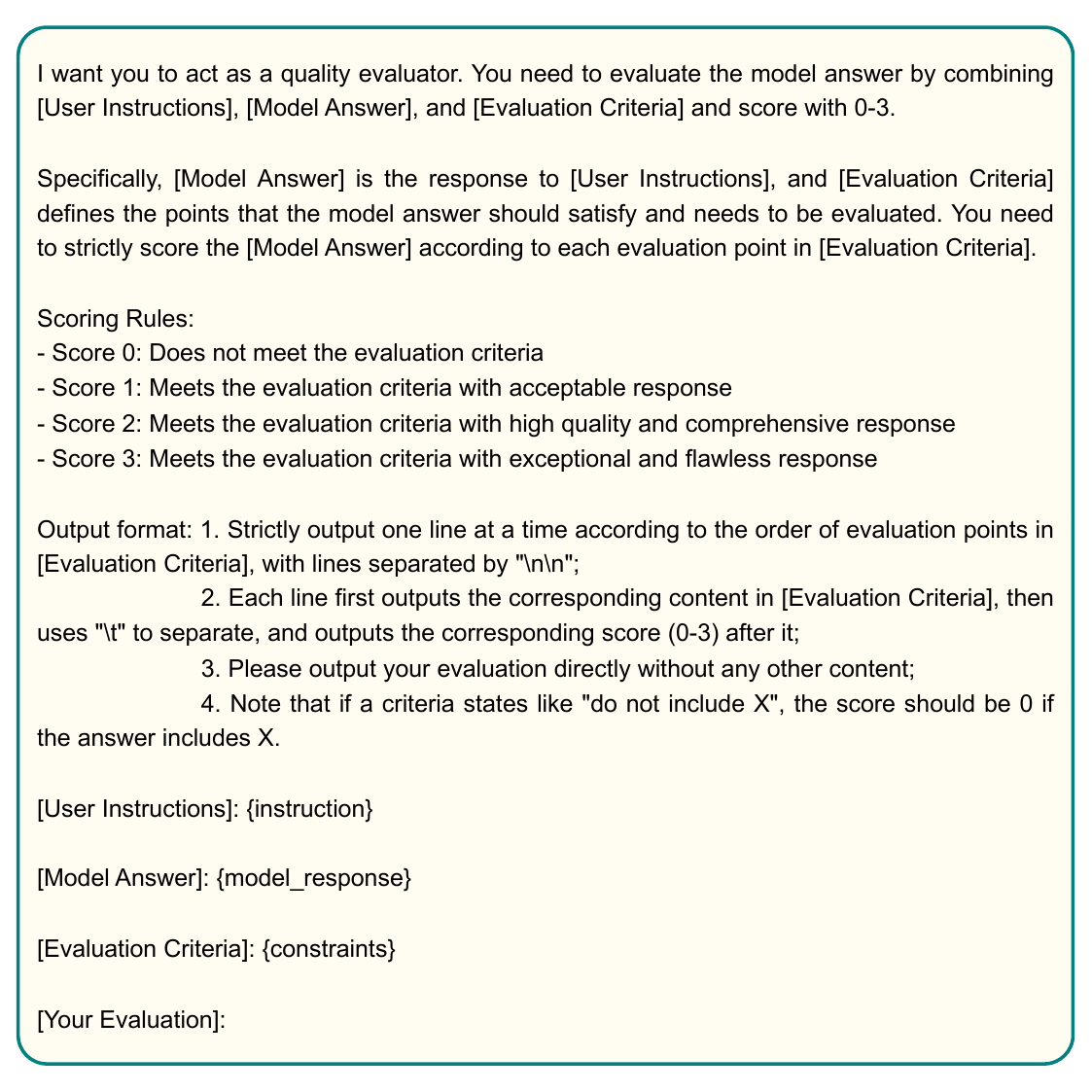}
\caption{Prompt for verifying model responses against constraints.}
\label{figure:cons_check}
\end{figure}

\begin{figure}[h]
\centering
\includegraphics[width=0.8\linewidth]{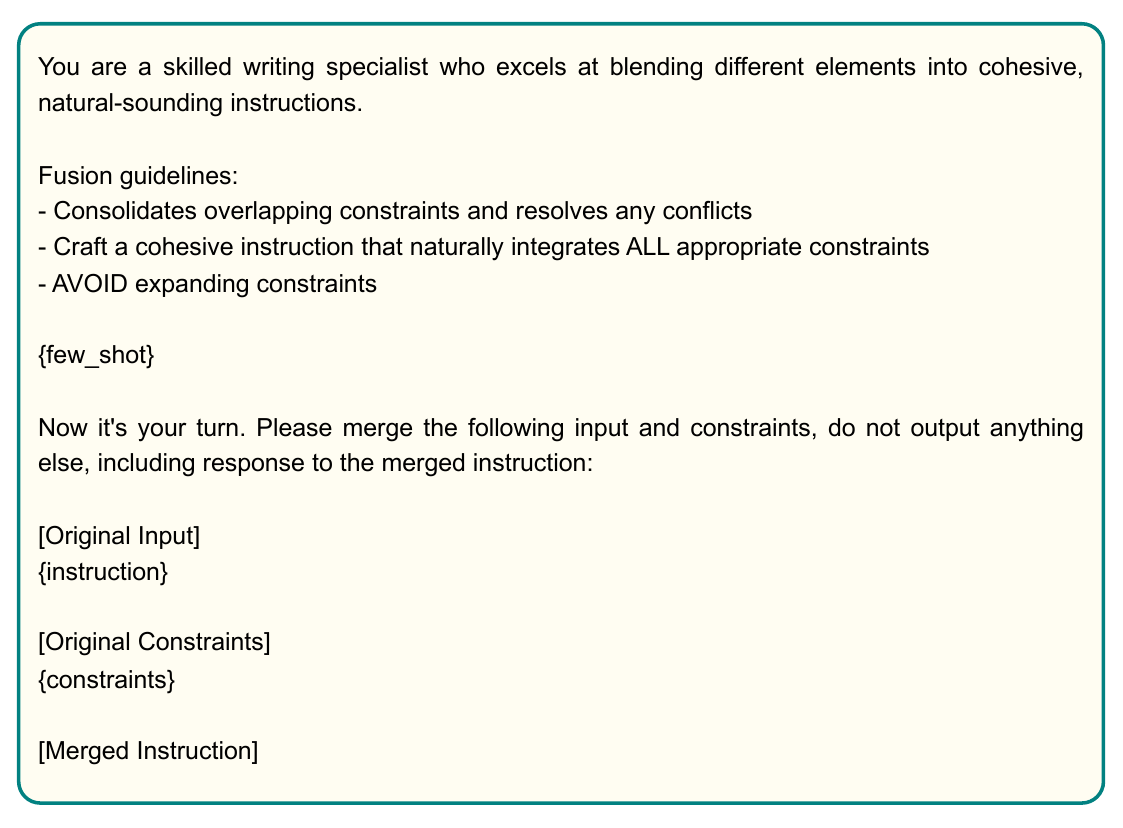}
\caption{Prompt for combining instructions with constraints.}
\label{figure:cons_combine}
\end{figure}

This section presents the prompts used in our data generation pipeline in Iterative Instruction Refinement step. These prompts serve different purposes in our methodology, from constraint generation (Figure \ref{figure:cons_generate}), constraint verification (Figure \ref{figure:cons_check}), and finally combines all elements into refined instructions (Figure \ref{figure:cons_combine}).

\section{Instruction Score Examples}
\label{appendix:ins_score_case}

\begin{figure}[h]
\centering
\includegraphics[width=0.95\linewidth]{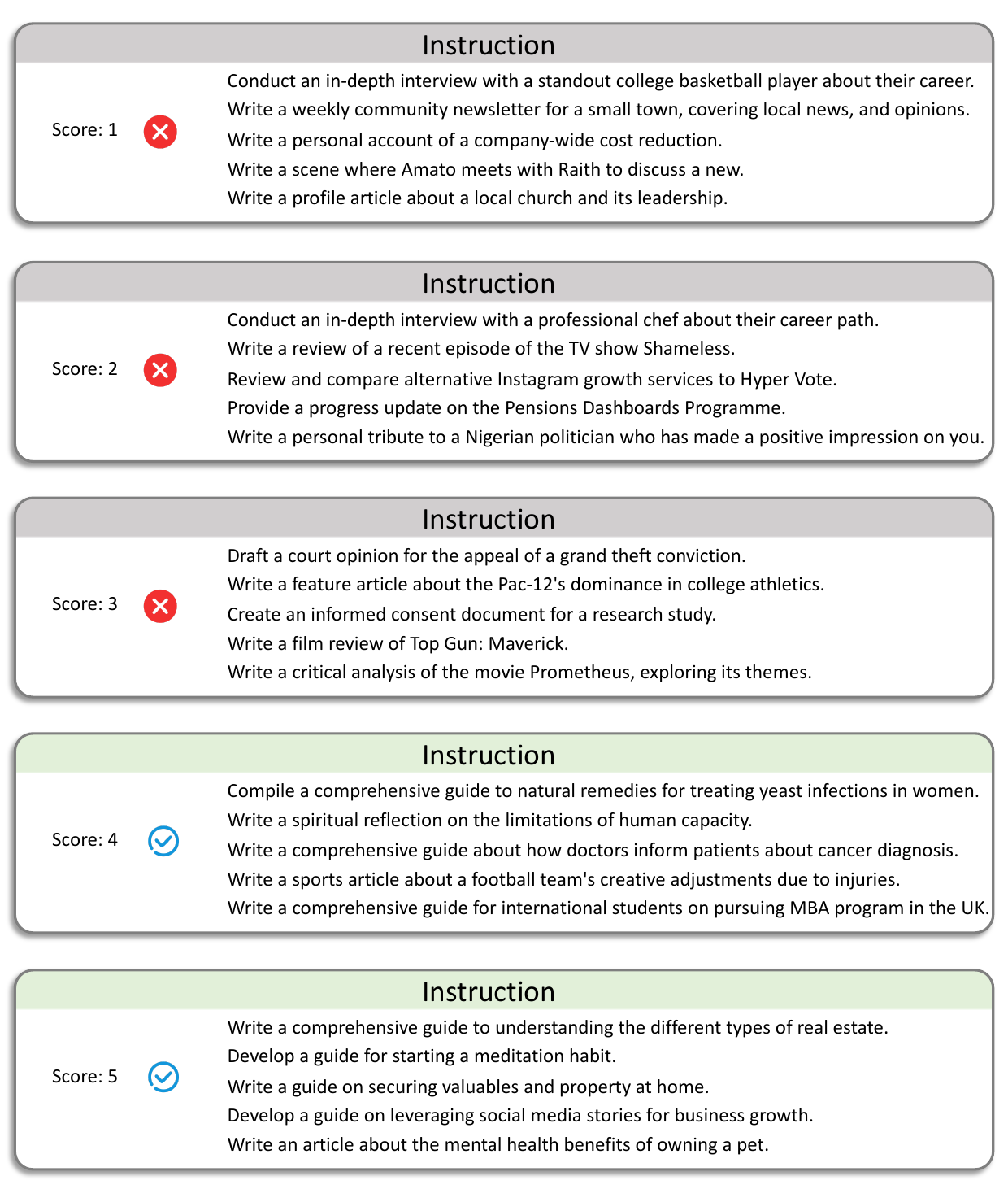}
\caption{Examples of instructions at different score levels (1-5), where each score level is illustrated with five representative cases. Score 1 represents the lowest quality while score 5 represents the highest quality.}
\label{figure:ins_score_case_show}
\end{figure}

This section presents a comprehensive analysis of instruction quality through representative examples. As illustrated in Figure \ref{figure:ins_score_case_show}, we provide a diverse set of instructions spanning the entire quality spectrum (scores 1-5). Each score category is exemplified by five carefully selected cases, where score 1 represents basic quality and score 5 demonstrates exceptional quality.


\label{sec:appendix}
\end{document}